\documentclass{article}

\usepackage{arxiv}

\usepackage[utf8]{inputenc} 
\usepackage[T1]{fontenc}    
\usepackage{hyperref}       
\usepackage{url}            
\usepackage{booktabs}       
\usepackage{amsfonts}       
\usepackage{nicefrac}       
\usepackage{microtype}      
\usepackage{lipsum}
\usepackage{graphicx}
\usepackage{amssymb}
\usepackage{commath}
\usepackage{wrapfig}
\usepackage{placeins}
\usepackage{float}
\usepackage[normalem]{ulem}

\title{Convolutional Feature Extraction and Neural Arithmetic Logic Units for Stock Prediction}

\author{
    Shangeth Rajaa\thanks{\href{https://shangeth.github.io/}{shangeth.github.io}} \\
  Department of Mathematics\\
  BITS Pilani Goa Campus\\
  Goa, India 403725 \\
  \texttt{f20160442@goa.bits-pilani.ac.in} \\
   \And
 Jajati Keshari Sahoo \\
  Department of Mathematics\\
  BITS Pilani Goa Campus\\
  Goa, India 403725 \\
  \texttt{jksahoo@goa.bits-pilani.ac.in} \\
}

\begin{document}
\maketitle

\begin{abstract}
Stock prediction is a topic undergoing intense study for many years. Finance experts and mathematicians have been working on a way to predict the future stock price so as to decide to buy the stock or sell it to make profit. Stock experts or economists,  usually analyze on the previous stock values using technical indicators, sentiment analysis etc to predict the future stock price.  In recent years, many researches have extensively used machine learning for predicting the stock behaviour. In this paper we propose data driven deep learning approach to predict the future stock value with the previous price with the feature extraction property of convolutional neural network and to use Neural Arithmetic Logic Units with it.

\end{abstract}


\section{INTRODUCTION}

A large number of people buy and sell stocks everyday in an aim to make maximum profit. Many mathematical methods and models have been developed which analyses the movement of the stock price. But its not sure if the future stock prices can actually be predicted due to its dependency on various factors and its dynamic nature. 
In recent years, machine learning and deep learning are being used in almost all the industries including finance. Machine learning in one way can be viewed as a function approximation(or a complex multiple dimensional curve fitting) for a given data. Machine learning can analyse and learn the complex multiple dimensional features of the data which humans cannot visualize or learn.
Although there are several mathematical models and techniques for stock prediction, this paper focuses on data driven machine learning approach with least knowledge in finance.The future stock price is to be predicted given the past prices. This paper tries to use and analyse the complex feature extraction ability of deep learning to learn the pattern of the stock price movement and predict the future price. 
\section{MACHINE LEARNING}
In recent times machine learning research in finance has been steadily increasing. There are generally 2 types of tasks in machine learning, classification and regression. Supervised machine learning regression model will be used for this stock prediction task.

\subsection{Classical Machine Learning Algorithms}
Classical machine learning algorithms are much more easier to interpret and understand than deep learning as we have a thorough understanding of underlying algorithms. These algorithms works better even on smaller data set and are computationally cheaper than deep learning techniques. Many researches have been done in predicting the stock price using classical machine learning algorithms. The author of \cite{svm1} has used Support Vector Machine (SVM) for financial forecasting and also did experimental analysis of parameters for SVM. Random forest techniques are also used in financial data, in \cite{treemodel}. Random forest, Naive bayes and support vector machine are used for classification the direction of movement of financial data.

\subsection{Deep Learning}
Although many machine learning algorithms exists and are successful, the evolution of deep learning marked a great milestone in the field of Artificial intelligence. The base work for deep learning started in 1940s, but it became more popular recently due to availability of more data and cheap computation devices. The performance of deep learning models increased exponentially every year and is projected to increase more. Image classification task is performed in \cite{annimage} using a Artificial Neural networks. After Neural Networks, many new models were invented to increase the performance of deep learning in images, videos and time series data such as text, voice, etc. Convolutional Neural Network \cite{alexnet} won the imagenet competition as it was good in extracting features of images/frames. Then Recurrent Neural networks \cite{rnn} were used for series data such as text and voice which needed a memory to remember the previous data features. Deep learning also performs very good in unsupervised models such as Auto Encoder \cite{ae} , General Adverserial Networks(GAN) \cite{gan} and in Reinforcement Learning.


\section{Deep Learning in Finance}

\subsection{Artificial Neural Networks(ANN)} 
ANNs are models comprised of densely connected computation nodes(neurons). These neural networks have the ability to learn complex features of the input data and perform the task. ANNs are series of matrix multiplication with non-linear function to make the whole network non linear to learn more complex features. 
\vspace{-0.25cm}
\begin{equation} \label{eq:1}
h_1 = \phi(X.W_1 + b_1) 
\end{equation}
\begin{equation} \label{eq:2}
h_i = \phi(h_{i-1}.W_i + b_i) 
\end{equation}
\begin{equation} \label{eq:3}
\hat{y} =  \phi(h_{n}. W_n + b_n)
\end{equation}

where n is the number of layers in the network, h is the hidden unit , $\hat{y}$ is the prediction in forward pass through the model abd $\phi$ is the activation function.
\cite{ann1} and \cite{ann2} uses Artificial Neural Networks to predict the stock price and direction of movement of the price. Dimensionality reduction techniques such as Principle Component Analysis(PCA) are used in \cite{drann} for stock prediction.
Artificial neural networks are also experimented for the task of predicting close price after 5 time interval(days/hour/minute). Data got from data processing steps explained in PROPOSED APPROACH was used and Tensors of shape (n, 20) was used as input data , where n is the number of data. And tensor of shape (n,1) was the label. The model consists of 4 layers of Fully Connected Dense Layer with dropouts and ReLU Non Linearity.

\subsection{Convolutional Neural Network}
Convolutional Neural Network(CNN)s are stacks of convolution operations between input which is passed through the network and filters(kernels) which extract the features of the input. The network is also activated with some activation function like ReLU for non linearity . The dimension of the layers are reduced with Pooling layers to reduce computation  and it can also be viewed as increasing the feature concentration.
\cite{convfin1} shows the potential of convolutional neural network for finance stock prediction. 1-d convolutional network \cite{convfin2} is also used to predict the stock movement as a classification model with 1 day close, open, high, low, volume data.
For this experiment, since the data is 1 dimensional , Conv1d(1 dimensional convolutional layers) of Pytorch is used with 3 convolutional layers with MaxPooling and ReLU activation. Then the convolutional layers are flattened into tensor of shape (n, 1, -1), where n is the number of data in the batch and -1 represents length of the layer multiplied by number of channels in the last convolutional layer. Followed by 3 layers of Dense or Fully Connected Layers with ReLU activation and Dropouts to avoid over fitting of the data.

\subsection{Recurrent Neural Networks}
Recurrent Neural network predicts an output given an input but in a sequential manner. The inputs and outputs are in sequence like text or audio. 
\vspace{-0.25cm}
\begin{equation} \label{eq:4}
h_t = \phi(X_t . W_x + h_{t-1} . W_h)
\end{equation}
\begin{equation} \label{eq:3}
\hat{y_t} = \phi(h_t . W_y)
\end{equation}

where $W_x, W_y, W_y$ are the weights, $h_t$ is the hidden state or memory state of state/time t and $\phi$ is the activation function. 
The financial data can be seen as a sequential data , the future stock price is predicted in \cite{lstm1} using LSTM network. A hybrid model RNN was used in \cite{lstm2} to predict the stock price.

\subsection{Neural Arithmetic Logic Units}
Neural Networks, although can perform several tasks nearly to human level accuracy, but they seem to fail when it encounters quantities outside the range of training data, like extrapolation. This shows that that the models actually try to fit the data rather than to generalize and learn it.  \cite{nalu} proposed a new module Neural Accumulator and Neural Arithmetic Logic Units which can be added to any neural network architecture which helps in generalizing quantities to neural network and helps the model to generalize for tasks like extrapolation. 

Stock prediction in one way can also be seen as an extrapolation task , where we are trying to predict the stock price in the future which can be above or below the range of out training data. In this paper we propose to use the ability of the Neural Arithmetic Logic Units to generalize and extrapolate to our task of stock prediction.


\section{Proposed Approach}

\subsection{Data}
\vspace{-0.8cm}
\begin{figure}[ht]
\centering
\includegraphics[width=1.0\textwidth]{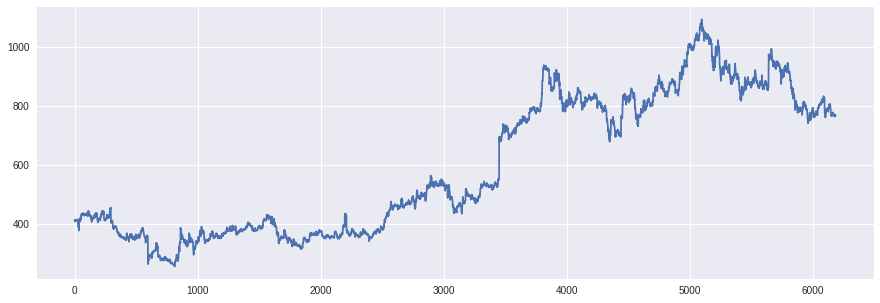}
      \caption{Closing Stock Prices data}
      \label{close_data}
\end{figure}

\vspace{-0.5cm}
Historical stock price data of India from Feb, 2015 to Aug, 2018 was used for this research. The data contains columns like Date, Close, High, Low, Open, Volume. This data changes every 1 hour, a total of around 6200 price data. The data set is checked for missing data and removed. Only Close prices are taken. All the other columns such as Date, High, Low, Open, Volume are omitted in the data. The goal is to predict stock closing price after 5 interval, with the closing price of past 20 intervals. This is a regression task to predict the exact closing price. For computational reasons and faster convergence, the data is scaled to a range of 0-1. The stock values are scaled with 
\begin{equation} \label{eq:3}
x_{scaled} = \dfrac{x-x_{min}}{x_{max}-x_{min}}
\end{equation}


\begin{table}[htbp]
\centering
\caption{Scaled Close Prices data}
\label{tab:scaled_price}
\begin{tabular}{|l|l|l|l|l|l|}
\hline
\textbf{Close Price}    & 411.15 & 414.05 & 410.20 & 410.25 & 410.00\\            \hline
\textbf{Scaled Close} & 0.1840 & 0.1874 & 0.1828& 0.1829& 0.1826\\          \hline
\end{tabular}
\end{table}


After scaling, the data is split into input and label. Input contains past 20 scaled close prices and the label contains the scaled stock prices after 5 intervals.

Facebook's PyTorch framework was used to design the computation graph and for training the model. The arrays of data are converted into tensors and are split into batches for faster computation using the advantage of Matrix operations.
So the input X will be a vector of shape (20, 1) and label will be of shape (1, 1).

The data was split into training and testing data in the ration of (8:2). And a batch size of 1232 was used to split the data into 5 equal batches. So 4 batches of 1232 data for training set and 1 batch for test set.
Each batch of data will be a tensor of shape (1232, 20) for Artificial Neural network models and tensor of shape (1232, 1, 20) for Convolutional Neural Network models.

\subsection{Neural Arithmetic Logic Units(NALU) based model for Stock Prediction}
Instead of PyTorch's nn.Linear layers, a self defined NALU module which is defined by

Neural Accumulator(NAC):\vspace{-0.25cm}
\begin{equation} \label{eq:3}
a = Wx
\end{equation}
\begin{equation} \label{eq:3}
W = \tanh(\hat{W})\circledast \sigma(\hat{M}) \\
\end{equation}

Neural Arithmetic Logic Unit(NALU):
\begin{equation} \label{eq:3}
y = g\circledast a + (1-g) \circledast m  
\end{equation}
\begin{equation} \label{eq:3}
m=\sigma(\ W(log(\abs{x} +\epsilon)))
\end{equation}
\begin{equation} \label{eq:3}
g=\sigma(Gx)
\end{equation}

Sigmoid function was used in the calculation of m instead of exponential function which was used originally in the Neural Arithmetic Logical units paper.
Four layers of Neural Arithmetic Logic Units are stacked like fully connected layers using defined pytorch NALU module. Dropouts are added in between each layer as a regularization technique to avoid over fitting the data. 
Relu activation function is added in between the NALU layers.

\vspace{-0.25cm}
\begin{equation} \label{eq:3}
ReLU(x) =\begin{cases}
   0 & x \leq 0\\    
   x & x >  0\\  
\end{cases}
\end{equation}

Finally sigmoid activation is used to make the prediction in the desired range of 0-1 (as the data is scaled to 0-1 range). 
\begin{equation} \label{eq:3}
Sigmoid(x) =  \dfrac{1}{1+e^{-x}}
\end{equation}


\begin{figure}[htbp]
\centering
\includegraphics[height=5cm, width=8cm]{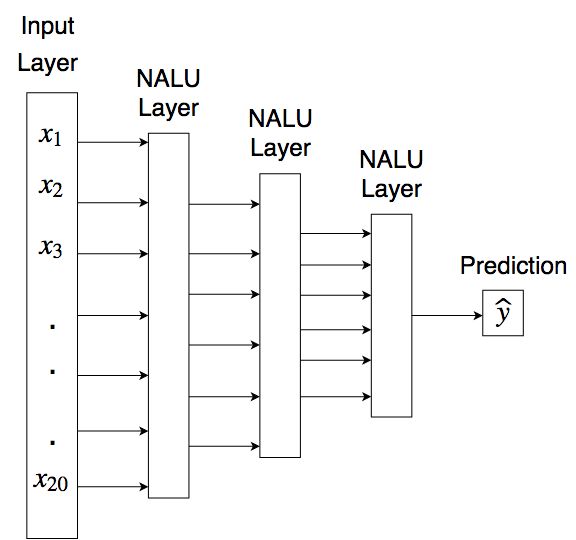}
\caption{Architecture of NALU Network}
\label{fig:naluarc}
\end{figure}

The output of the network is compared with the true value using Squared L2 Norm(Mean Squared Error) loss function.

\begin{equation} \label{eq:3}
MSELoss(y, \Hat{y}) = \dfrac{1}{m}\sum_{i=1}^{m}(y^{(i)} - \Hat{y}^{(i)})^2 
\end{equation}
where $ y^{(i)}$ is the true label value and $\hat{y}^{(i)} $ is the model prediction for $i^{th}$ training data.To minimize the loss, back propagation algorithm is used with Adam optimizer. A cyclic learning rate \cite{clr} scheduler has been used with the optimizer as an attempt to escape the problem of local minimum of loss. When the algorithm is stuck in a local minimum or narrow minimum , increasing the learning rate help it escape the local space and reach a better or wider minimum space. Each data batch is has been used 500 times to learn and update the weight parameters of the model so as to reduce the total loss. As we use cyclic learning rate, the loss tends to go high when the learning rate increases, so we save the model state with lowest loss.


\subsection{Convolutional feature extraction and NALU based model for Stock Prediction}
Convolutional Neural Network has been used to predict the stock in the past. This paper proposes a new model using the feature extraction ability of convolutional neural network with the Neural Arithmetic Logic Units. 
As the stock data is 1 dimensional series data, 1 dimensional convolutional layers using nn.Conv1d in Pytorch are used and stacked 3 layers of 1-d convolutional layers to extract the features of stock price movements. Kernel size of 4 has been used in the network for all the convolutional layers. The number of kernels/filters in each layers are 1, 16, 32 and 64. Max pooling layers are added in between every convolutional layer to reduce the dimension , kernel size of 1 or 2 is used and stride is also 2 , which will reduce the layer length to half. ReLU activation function is used to make the network non linear.

\begin{figure}[htbp]
\centering
\includegraphics[height=6cm, width=10cm]{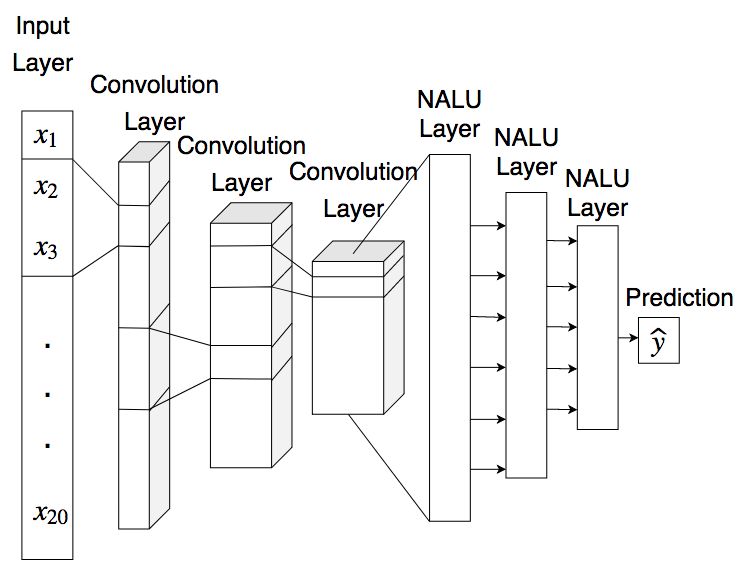}
\caption{Architecture of CNN-NALU Network}
\label{fig:cnnnaluarc}
\end{figure}
Convolutional layers are followed by 2 layers of Neural Arithmetic Logic Units and 2 layers of Fully connected layers as the regressor. ReLU activation function is used in between the linear and NALU layers with dropouts to avoid overfitting of the data. we use sigmoid activation function in the last layer of the network to make the prediction in the range of 0-1 .
Squared L2 Norm loss function was used to get the loss after the forward pass, Adam optimizer was used for optimization and Cyclic learning rate scheduler was used to change the learning rate in cycle from $10^{-6}$ to $10^{-2}$.  



\section{Results}
Different models were used in this research to find which model is able to learn the trend of the stock price and predict the future price given the last 20 prices better.
In each iteration after training the models using training set, the testing set is used to check how good the model has learned and how good it can predict unseen data. After training the model, the whole stock close data is predicted using the trained model and plotted to visualize how good the model performs on the data as a whole.

TABLE 2 gives the training loss of each of the model. It can be observed that Models with Neural Arithmetic Logic Units learned better ANN and CNN models.TABLE 3 gives the loss of the models on testing set. Models with Neural Arithmetic Logic Units was able to predict the stock price better than ANNs and CNNs on unseen data.After the training and validating the testing set, the model was used to test the complete data . Previous 20 data points were given and the model predicted the close price after 5 intervals. The loss of the model with the whole data set is given by TABLE 4. This value has to be re scaled back to the original interval to compare with the actual price. 

\begin{table}[htbp]
\centering
\caption{Training Loss of Models}
\label{tab:loss1}
\begin{tabular}{|l|l|}
\hline
\textbf{Model}                              & \textbf{Training Loss} \\ \hline
Artificial Neural Network(ANN)              & 8.04649e-06            \\ \hline
Convolutional Neural Network(CNN)           & 5.58822e-06  \\ \hline
Neural Arithmetic Logic Units Network(NALU) & 1.91356e-06            \\ \hline
NALU CNN Network(NALU-CNN)                  & 5.58499e-07            \\ \hline
\end{tabular}
\end{table}

\begin{table}[htbp]
\centering
\caption{Testing loss of Models}
\label{tab:loss2}
\begin{tabular}{|l|l|}
\hline
\textbf{Model}                              & \textbf{Testing Loss} \\ \hline
Artificial Neural Network(ANN)              & 1.30709e-06           \\ \hline
Convolutional Neural Network(CNN)           &  5.99638e-07         \\ \hline
Neural Arithmetic Logic Units Network(NALU) &  4.31875e-07         \\ \hline
NALU CNN Network(NALU-CNN)                  & 3.05196e-07           \\ \hline
\end{tabular}
\end{table}

\begin{table}[htbp]
\centering
\caption{Loss of Models in the whole data set}
\label{tab:loss3}
\begin{tabular}{|l|l|}
\hline
\textbf{Model}                              & \textbf{Total Loss} \\ \hline
Artificial Neural Network(ANN)              & 1.29998e-06         \\ \hline
Convolutional Neural Network(CNN)           & 1.07971e-06          \\ \hline
Neural Arithmetic Logic Units Network(NALU) & 3.97540e-07        \\ \hline
NALU CNN Network(NALU-CNN)                  & 3.30627e-07         \\ \hline
\end{tabular}
\end{table}

\begin{figure*}[htbp]
  \includegraphics[height=6cm,width=\textwidth]{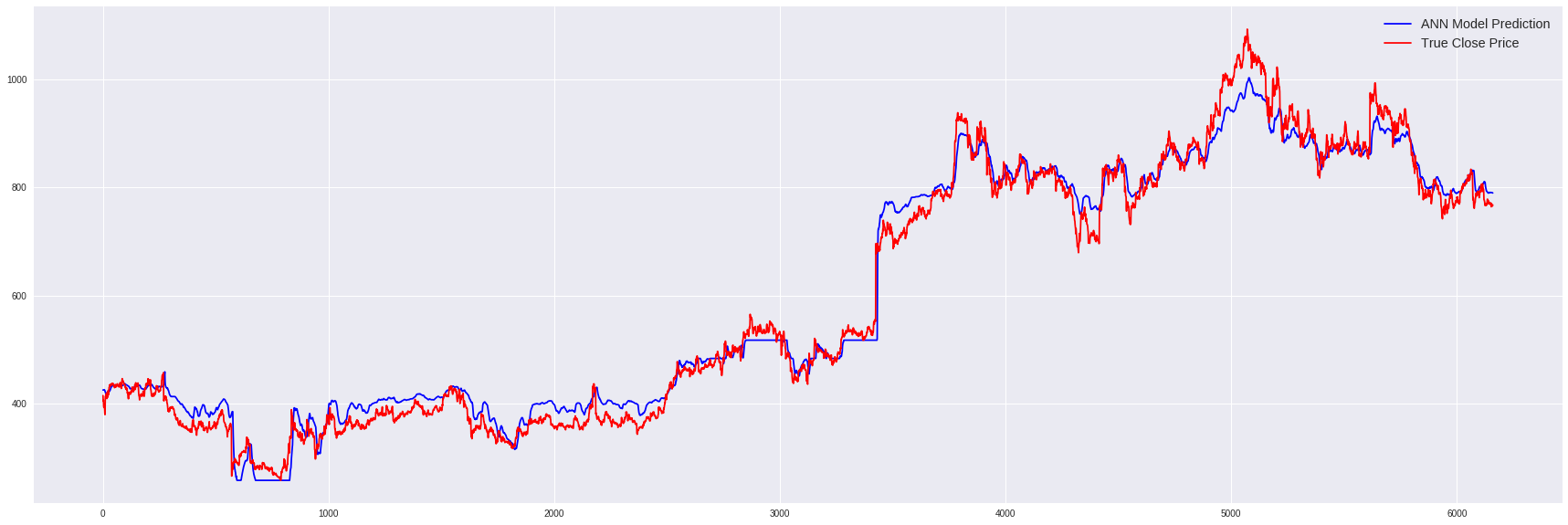}
  \caption{Prediction plot of Artificial Neural Network Model}
\end{figure*}

\begin{figure*}[htbp]
  \includegraphics[height=6cm,width=\textwidth]{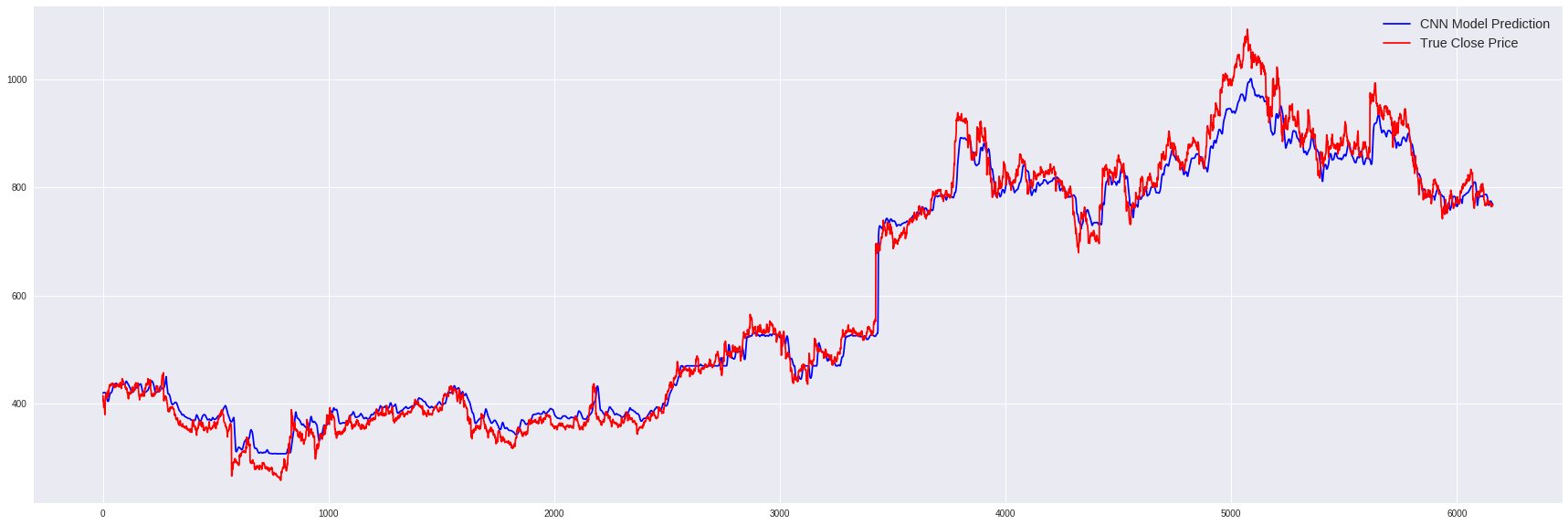}
  \caption{Prediction plot of Convolutional Neural Network Model}
\end{figure*}

\begin{figure*}[htbp]
  \includegraphics[height=6cm,width=\textwidth]{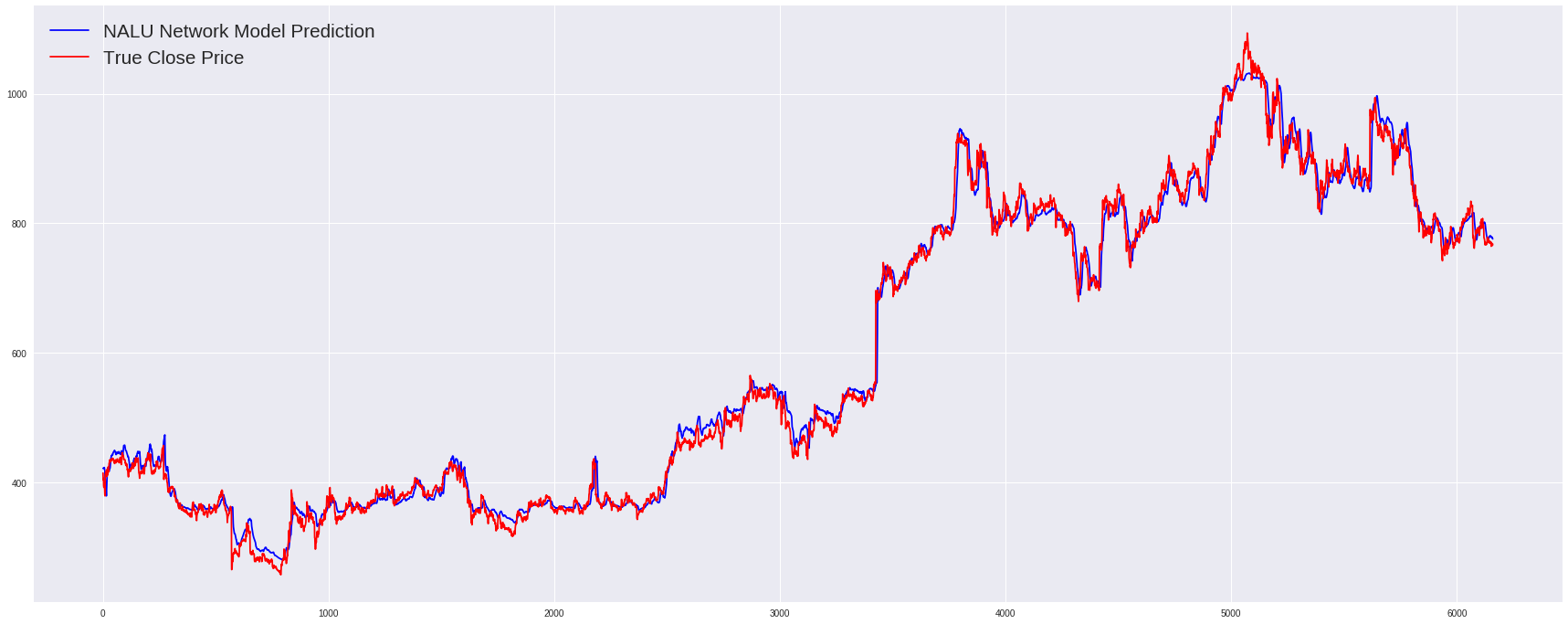}
  \caption{Prediction plot of NALU Network Model}
\end{figure*}

\begin{figure*}[htbp]
  \includegraphics[height=6cm,width=\textwidth]{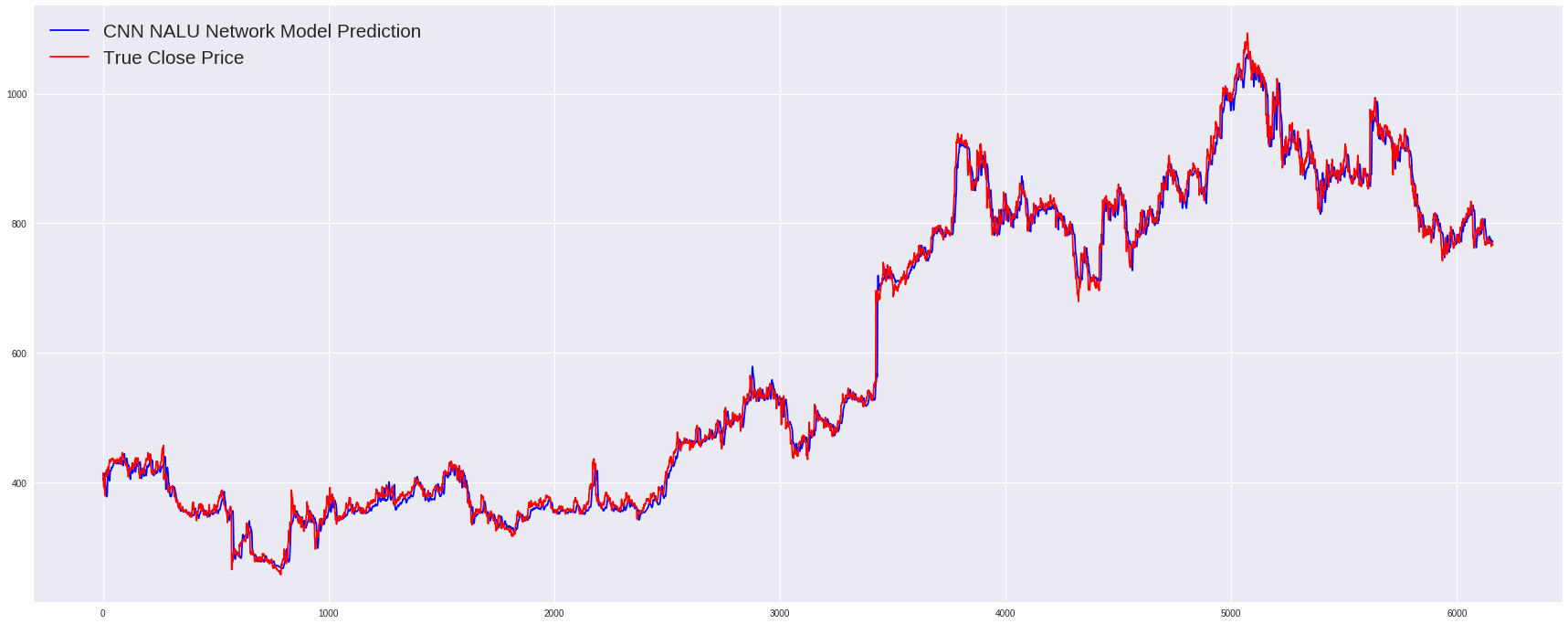}
  \caption{Prediction plot of CNN-NALU Network Model}
\end{figure*}

\section{Conclusion}
In this paper we proposed to use the feature extraction property of convolutional neural networks and the extrapolation and arithmetic ability of Neural Arithmetic Logic Units to predict the stock price 5 days later. 

During the course of this experiment it was observed that the models with Neural Arithmetic Logic Units(NALU) converged faster than the other model not only in the task of Stock prediction but also on many other tasks. NALU models were able to learn the pattern and other features of the stock values and was able to predict the closing price better than ANNs and CNNs.

\bibliographystyle{unsrt}
\bibliography{paper}

\begin{thebibliography}{10}

\bibitem{svm1}
L.~J. Cao and F.~E.~H. Tay.
\newblock Support vector machine with adaptive parameters in financial time
  series forecasting.
\newblock {\em IEEE Transactions on Neural Networks}, 14(6):1506--1518, Nov
  2003.

\bibitem{treemodel}
Jigar Patel, Sahil Shah, Priyank Thakkar, and K~Kotecha.
\newblock Predicting stock and stock price index movement using trend
  deterministic data preparation and machine learning techniques.
\newblock {\em Expert Systems with Applications}, 42(1):259 -- 268, 2015.

\bibitem{annimage}
Prof S~K Shah.
\newblock Image classification based on textural features using artificial
  neural network (ann).

\bibitem{alexnet}
Alex Krizhevsky, Ilya Sutskever, and Geoffrey~E Hinton.
\newblock Imagenet classification with deep convolutional neural networks.
\newblock In F.~Pereira, C.~J.~C. Burges, L.~Bottou, and K.~Q. Weinberger,
  editors, {\em Advances in Neural Information Processing Systems 25}, pages
  1097--1105. Curran Associates, Inc., 2012.

\bibitem{rnn}
J.~T. Connor, R.~D. Martin, and L.~E. Atlas.
\newblock Recurrent neural networks and robust time series prediction.
\newblock {\em IEEE Transactions on Neural Networks}, 5(2):240--254, March
  1994.

\bibitem{ae}
Pascal Vincent, Hugo Larochelle, Yoshua Bengio, and Pierre-Antoine Manzagol.
\newblock Extracting and composing robust features with denoising autoencoders.
\newblock In {\em Proceedings of the 25th International Conference on Machine
  Learning}, ICML '08, pages 1096--1103, New York, NY, USA, 2008. ACM.

\bibitem{gan}
Ian Goodfellow, Jean Pouget-Abadie, Mehdi Mirza, Bing Xu, David Warde-Farley,
  Sherjil Ozair, Aaron Courville, and Yoshua Bengio.
\newblock Generative adversarial nets.
\newblock In Z.~Ghahramani, M.~Welling, C.~Cortes, N.~D. Lawrence, and K.~Q.
  Weinberger, editors, {\em Advances in Neural Information Processing Systems
  27}, pages 2672--2680. Curran Associates, Inc., 2014.

\bibitem{ann1}
Yakup Kara, Melek~Acar Boyacioglu, and Ömer Kaan~Baykan.
\newblock Predicting direction of stock price index movement using artificial
  neural networks and support vector machines: The sample of the istanbul stock
  exchange.
\newblock {\em Expert Systems with Applications}, 38(5):5311 -- 5319, 2011.

\bibitem{ann2}
K.~Abhishek, A.~Khairwa, T.~Pratap, and S.~Prakash.
\newblock A stock market prediction model using artificial neural network.
\newblock In {\em 2012 Third International Conference on Computing,
  Communication and Networking Technologies (ICCCNT'12)}, pages 1--5, July
  2012.

\bibitem{drann}
Chih-Fong Tsai and Yu-Chieh Hsiao.
\newblock Combining multiple feature selection methods for stock prediction:
  Union, intersection, and multi-intersection approaches.
\newblock {\em Decision Support Systems}, 50(1):258 -- 269, 2010.

\bibitem{convfin1}
J.~Chen, W.~Chen, C.~Huang, S.~Huang, and A.~Chen.
\newblock Financial time-series data analysis using deep convolutional neural
  networks.
\newblock In {\em 2016 7th International Conference on Cloud Computing and Big
  Data (CCBD)}, pages 87--92, Nov 2016.

\bibitem{convfin2}
Sheng Chen and Hongxiang He.
\newblock Stock prediction using convolutional neural network.
\newblock {\em IOP Conference Series: Materials Science and Engineering},
  435(1):012026, 2018.

\bibitem{lstm1}
Kai Chen, Yi~Zhou, and Fangyan Dai.
\newblock A lstm-based method for stock returns prediction: A case study of
  china stock market.
\newblock In {\em Proceedings of the 2015 IEEE International Conference on Big
  Data (Big Data)}, BIG DATA '15, pages 2823--2824, Washington, DC, USA, 2015.
  IEEE Computer Society.

\bibitem{lstm2}
Akhter~Mohiuddin Rather, Arun Agarwal, and V.N. Sastry.
\newblock Recurrent neural network and a hybrid model for prediction of stock
  returns.
\newblock {\em Expert Systems with Applications}, 42(6):3234 -- 3241, 2015.

\bibitem{nalu}
Andrew Trask, Felix Hill, Scott~E Reed, Jack Rae, Chris Dyer, and Phil Blunsom.
\newblock Neural arithmetic logic units.
\newblock In S.~Bengio, H.~Wallach, H.~Larochelle, K.~Grauman, N.~Cesa-Bianchi,
  and R.~Garnett, editors, {\em Advances in Neural Information Processing
  Systems 31}, pages 8046--8055. Curran Associates, Inc., 2018.

\bibitem{clr}
Leslie~N. Smith.
\newblock No more pesky learning rate guessing games.
\newblock {\em CoRR}, abs/1506.01186, 2015.

\end{thebibliography}
\end{document}